\documentclass[11pt]{article}

\usepackage[preprint]{acl}

\usepackage{times}
\usepackage{latexsym}

\usepackage[T1]{fontenc}

\usepackage[utf8]{inputenc}

\usepackage{microtype}

\usepackage{inconsolata}

\usepackage{graphicx}
\usepackage{booktabs}
\usepackage{amsmath}
\usepackage{amsfonts}
\usepackage{multirow}
\usepackage{xspace}
\usepackage{enumitem}

\newcommand{\sysname}{\texttt{SOPRAG}\xspace}

\usepackage[most]{tcolorbox}

\newtcolorbox{promptbox}[1]{
    colback=gray!5!white,      
    colframe=gray!75!black,    
    coltitle=black,            
    fonttitle=\bfseries,       
    fontupper=\small,
    title=#1,                  
    enhanced,
    attach title to upper,     
    after title={\par\medskip\hrule\medskip}, 
    arc=2pt,                   
    boxrule=0.5pt,             
    left=5pt, right=5pt, top=5pt, bottom=5pt
}

\title{\sysname: Multi-view Graph Experts Retrieval for Industrial Standard Operating Procedures}

\author{Liangtao Lin, \quad
        Zhaomeng Zhu, \quad
        Tianwei Zhang \and Yonggang Wen \\
    Nanyang Technological University, Singapore \\
    liangtao002@e.ntu.edu.sg, \{tianwei.zhang, ygwen\}@ntu.edu.sg}

\begin{document}
\maketitle

\begin{abstract}

Standard Operating Procedures (SOPs) are essential for ensuring operational safety and consistency in industrial environments. However, retrieving and following these procedures presents unique challenges, such as rigid proprietary structures, condition-dependent relevance, and actionable execution requirement, which standard semantic-driven Retrieval-Augmented Generation (RAG) paradigms fail to address. 
Inspired by the Mixture-of-Experts (MoE) paradigm, we propose \textbf{\sysname}, a novel framework specifically designed to address the above pain points in SOP retrieval. \sysname replaces flat chunking with specialized Entity, Causal, and Flow graph experts to resolve industrial structural and logical complexities. To optimize and coordinate these experts, we propose a Procedure Card layer that prunes the search space to eliminate computational noise, and an LLM-Guided gating mechanism that dynamically weights these experts to align retrieval with operator intent.
To address the scarcity of domain-specific data, we also introduce an automated, multi-agent workflow for benchmark construction. Extensive experiments across four industrial domains demonstrate that \sysname significantly outperforms strong lexical, dense, and graph-based RAG baselines in both retrieval accuracy and response utility, achieving perfect execution scores in real-world critical tasks.
\end{abstract}

\section{Introduction}
\label{sec:intro}
The Standard Operating Procedure (SOP) is a set of step-by-step instructions compiled by organizations to guide personnel in performing routine task, which is essential for ensuring safe, consistent, and efficient operations in industrial environments. As industrial systems grow increasingly complex, \textit{SOP Retrieval}, the capability to assist operators to rapidly identify and follow the appropriate procedure, has become ever more critical for minimizing downtime and mitigating operational risks~\cite{akyar2012standard}.

However, retrieving SOPs presents unique challenges compared to the general open-domain document retrieval. Unlike common corpora structured around narrative semantics, SOPs are engineered for operational execution, characterized by rigid dependencies between critical entities, conditional logic, and procedural workflows. This distinct nature renders standard semantic matching insufficient and imposes three primary hurdles for existing methods:

\begin{figure}[t]
    \centering
    \includegraphics[width=0.85\linewidth]{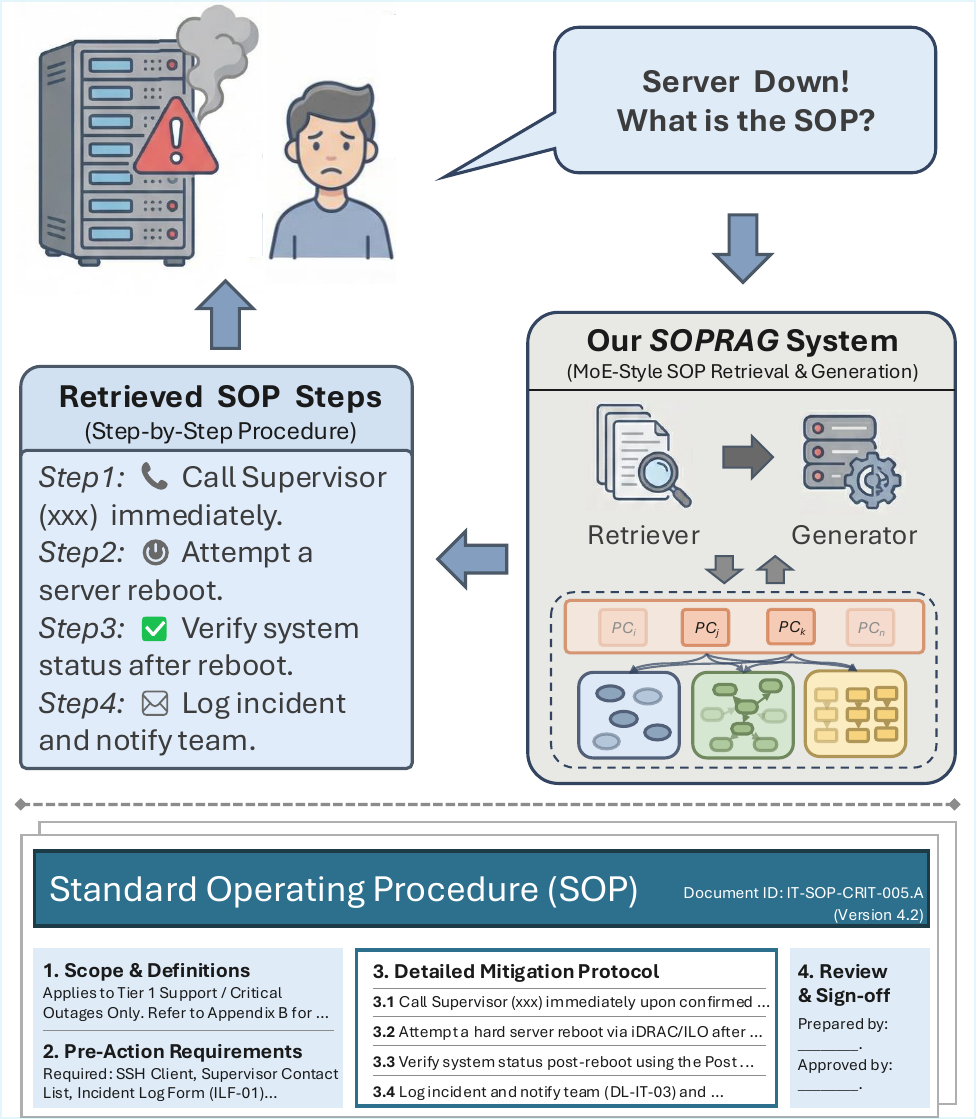}
    \caption{\textbf{Overview of SOP Retrieval in practice.} (Top) An operator encounters a system failure and uses our \sysname to generate executable step-by-step guidance. (Bottom) A representative sample of a SOP source document used for retrieval.}
    \label{fig:concept}
\end{figure}

\begin{itemize}[topsep=0pt, leftmargin=*, nosep]
    \item \textbf{(C1) Proprietary Structure:} SOPs possess a rigid hierarchy with high-level goals (titles), specific constraints (parameters, alarm codes, equipment models), and execution logic (steps). Traditional chunking strategies often sever semantic links between a procedural step and its governing context, rendering the retrieved chunk ambiguous or actionable only for the wrong equipment.
    \item \textbf{(C2) Condition-Dependent Relevance:} Operator queries are inherently scenario-specific. A symptom like "system overheat" may correspond to entirely different procedures depending on the underlying cause or system state. Relevance is determined by causal logic and intent recognition, rather than simple lexical or semantic overlap.
    \item \textbf{(C3) Actionable Response Generation:} Unlike general Question Answering (QA), where a factual summary suffices, industrial operators require executable knowledge. The desired output is a precise, step-by-step workflow that guides operators through a specific scenario without hallucinations that may lead to serious consequences.
\end{itemize}

Current Retrieval-Augmented Generation (RAG) paradigms fall short in addressing these challenges. While GraphRAG approaches \cite{graphragsurvey} have improved models' reasoning capabilities, they primarily construct graphs based on semantic similarity or entity co-occurrence. They often neglect the causal and sequential logic that defines a workflow, leading to suboptimal search quality and responses without procedural coherence. Furthermore, research in this niche is hindered by the lack of comprehensive benchmarks, as most existing datasets focus on general QA rather than the nuanced requirements of industrial retrieval.

Inspired by the Mixture-of-Experts (MoE) paradigm~\cite{Shazeer2017OutrageouslyLNmoe}, we propose \textbf{\sysname}, a structure-aware framework to bridge the above gaps by explicitly modeling the procedural "shape" of industrial knowledge. To dismantle the core hurdles of proprietary structure (C1), condition-dependency (C2), and actionable execution (C3), we introduce three specialized graph experts, \textbf{Entity, Causal, and Flow graphs}, each engineered to resolve specific structural and logical complexities. By utilizing graph-based search and modeling instead of flat semantic chunking, \sysname directly overcomes context fragmentation while significantly elevating reasoning transparency and precision.

Consistent with the principle of sparse activation, we design a \textbf{Procedure Card} layer inspired by the cognitive workflow of human experts who scan high-level titles to prune irrelevant information before drilling into technical details. This layer serves to purge computational noise and ensure that the underlying specialists operate with maximum focus. We further implement an intention-aware \textbf{LLM Router} as our gating mechanism, which dynamically modulates attention weights assigned to each expert based on the operator's specific query intent. This elegant alignment reveals a robust, gated system tailored for the high-precision demands of industrial SOP retrieval.

Furthermore, to address the scarcity of domain-specific data and rigorously evaluate our framework, we introduce an agent-based simulation environment for automated benchmark construction, which deploys a team of specialized agents, the \textit{Collector}, \textit{Archivist}, \textit{Auditor}, and \textit{Examiner}, to emulate the rigorous document processing lifecycle of human experts.

Our main contributions are as follows:
\begin{itemize}[topsep=0pt, leftmargin=*, nosep]
    \item We identify the specific challenges of SOP retrieval and propose a novel agnentic framework for its benchmark construction, which could facilitate future research.
    \item We introduce \textbf{\sysname}, a MoE-inspired structure-aware retrieval and generation framework specifically tailored for SOPs, which integrates a Procedure Card layer with Multi-view Graph Experts to address the unique challenges in SOP retrieval.
    \item Extensive experiments demonstrate that our approach significantly outperforms strong lexical, dense, and graph-based RAG baselines in both retrieval accuracy and response utility.
\end{itemize}

\section{Related Work}
\subsection{Procedural Question Answering}

Procedural QA examines how to answer questions grounded in stepwise texts. Early work by \citet{banerjee-bandyopadhyay-2012-question} formalized question types and answering strategies for procedural text, while \citet{Antoine2018iclr-Simu} modeled action dynamics and state changes using Neural Process Networks. Subsequent studies enhanced procedural understanding by incorporating external knowledge \citep{zhang2021Know} or structural representations, such as graph-guided QA generation \citep{pham-etal-2024-graph}. Procedural QA has also been extended to multimodal settings, including RecipeQA~\citep{yagcioglu-etal-2018-recipeqa} and ProMQA~\citep{hasegawa-etal-2025-promqa}, as well as interactive, program-guided answering \citep{Maitra2020Enabling}. 

However, few studies have targeted domain-critical procedural texts such as SOPs. Existing work includes QA models for human–machine interaction in manufacturing~\citep{RUIZ2023103988human–machine}, domain-centric RAG for smart manufacturing ~\citep{zhang2021Know}, or private knowledge-based LLMs for enterprise SOP scenarios~\citep{Xie2024Reserch}. Nevertheless, these studies mainly focus on question answering, whereas real industrial operations require retrieving and following complete SOP documents, which is the central gap our work aims to address.


\subsection{LLM Agents Enhanced by SOP}
Recent work has increasingly explored incorporating SOPs into LLM-based agents to improve controllability, reliability, and domain alignment. SOP-Agent~\citep{ye2025sop} proposes to empower general-purpose agents with domain-specific SOPs as explicit behavioral guidance, while SOP-Bench~\citep{nandi2025sop} introduces a benchmark of complex industrial SOPs to evaluate agents’ procedural understanding and execution capabilities. Along this, Agent-S~\cite{kulkarni2025agent} formulates SOPs as agentic workflows, enabling LLM agents to automate real-world operational procedures. 

Other studies integrate SOPs into agent planning and coordination: ChatSOP~\cite{li-etal-2025-chatsop} employs SOP-guided Monte Carlo Tree Search for controllable dialogue planning, and Flow-of-Action~\citep{Pei2025Flow-of-Action} enhances multi-agent systems with SOP constraints for root cause analysis. Some efforts investigate SOP structuring and interpretation, including generating structured plan representations from procedural text~\cite{garg2025generating} and classifying SOP step components via prompt engineering~\citep{Bashatah2024promptSOP}. 

Overall, these works treat SOPs primarily as guidance or control mechanisms for LLM agents, rather than as retrievable procedural knowledge, leaving open the challenge of accurately retrieving and grounding agents in complete SOP documents in industrial settings.


\subsection{Graph-based RAG}
Retrieval-Augmented Generation enhances LLMs by grounding generation in retrieved external documents, but conventional flat chunk-based retrieval strategy ignores relationships across text segments. Graph-based RAG~\citep{graphragsurvey} addresses this limitation by organizing documents as graphs, enabling multi-hop retrieval, global context aggregation, and structured reasoning. Representative approaches include GraphRAG, where \citet{edge2024local} construct document graphs and perform community-level summarisation for query-focused global reasoning, and LightRAG~\citep{guo-etal-2025-lightrag}, which introduces a lightweight graph indexing scheme to jointly retrieve fine-grained evidence and high-level concepts. 

Beyond text graphs, \citet{Sun2023ThinkonGraphDA} propose Think-on-Graph, enabling LLMs to traverse knowledge graphs for multi-hop reasoning. \citet{li-etal-2024-graphreader} introduce GraphReader, which organizes long documents as graphs to support agent-style coarse-to-fine retrieval. More recently, \citet{gutiérrez2024hipporag,gutiérrez2025ragmemorynonparametriccontinual} move from RAG to non-parametric memory by structuring retrieved knowledge as persistent graphs for continual learning. Despite these advances, existing Graph-based RAG methods remain largely semantic-driven and text-centric, and do not explicitly model the procedural, hierarchical, and step-dependent structures of SOP documents. This motivates our structure-aware Graph-based RAG design for industrial SOP.


\section{SOPRAG}
\label{sec:methodology}

\subsection{Overview}
\begin{figure*}[ht]
    \centering
    \includegraphics[width=0.73\linewidth]{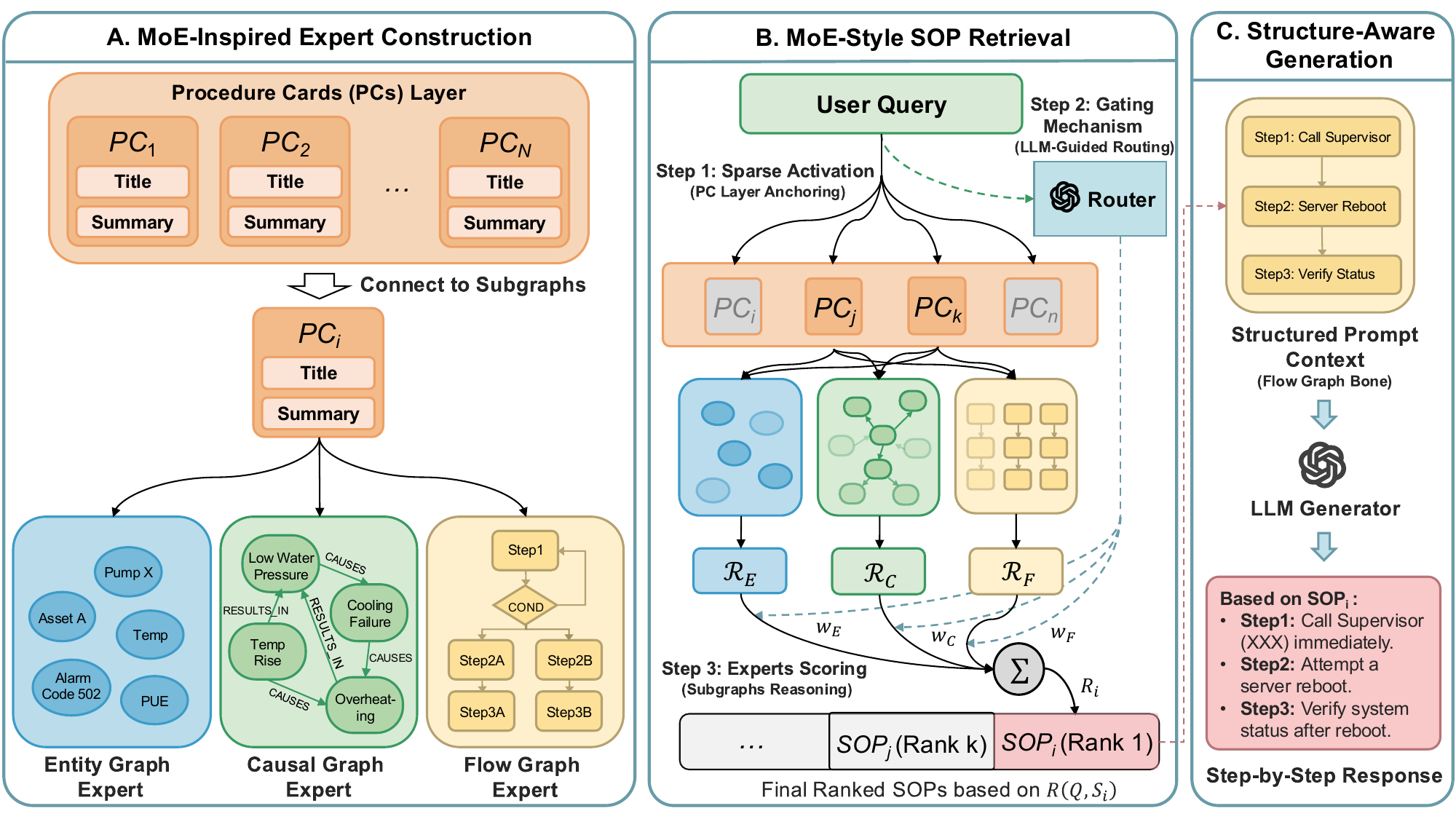}
    \caption{\textbf{Architecture overview of \sysname.} (A) Expert Construction: delineating the Procedure Card (PC) layer and the three multi-view graph experts (Entity, Causal, and Flow); (B) MoE-Style Retrieval: illustrating the sparse activation of the PC layer, the LLM-guided gating mechanism (Router), and the scoring and aggregation process of the graph experts; (C) Structure-Aware Generation: showing the generation of a step-by-step actionable response using the flow graph of retrieved SOP as a structured prompt context. (Further extensions are shown in Appendix~\ref{app:agent-exe}.)}
    \label{fig:soprag}
\end{figure*}

To address the unique challenges of SOP retrieval (C1, C2, and C3), while simultaneously overcoming the length and semantic constraints of chunking and structural deficiencies in existing graph-based RAG paradigms, we propose \textbf{\sysname}, as shown in Figure \ref{fig:soprag}. Inspired by the MoE paradigm, we introduce three specialized graph-based experts, the \textbf{Entity, Causal, and Flow graphs}, each specifically engineered to dismantle these industrial hurdles. To optimize this process, we introduce a \textbf{Procedure Card (PC)} layer as a Sparse Activation layer to prune the reasoning scope and reduce computational noise. This design is inspired by the cognitive workflow of human experts who prioritize scanning high-level titles to filter candidate procedures before drilling into technical execution. Finally, an \textbf{LLM-Guided Router} is employed for intent recognition, acting as a gating mechanism that dynamically assigns attention weights to the graph experts, ensuring the retrieval process is adaptively tailored to the user's specific query.

\subsection{Specialized Expert Construction}
Let $\mathcal{S} = \{S_1, S_2, \ldots, S_N\}$ be the set of atomic SOPs. We construct a knowledge base $\mathcal{G}$ where each SOP is represented through a sparse activation anchor and three specialized reasoning experts.

\subsubsection{Procedure Card Layer}
Consistent with the principle of sparse activation in MoE, we abstract each SOP $S_i$ into a \textit{Procedure Card} ($PC_i$), which consists of the SOP's high-level title and an LLM-synthesized one-line abstract:
\begin{equation}
    PC_i = \{ \text{Title}(S_i), \text{Abstract}(S_i) \} \in \mathcal{V}_{Global}
\end{equation}

The PC layer acts as the primary entry point, selecting only the relevant "expert clusters" to ensure the underlying reasoning components operate with maximum focus.

\subsubsection{Multi-view Graph Experts}
To resolve the specific hurdles of SOPs, we replace generic semantic chunks with three specialized graph experts, each engineered to process a distinct dimension of procedural information.

\paragraph{Entity Graph.}
To resolve the ambiguity of Proprietary Structure (C1), where procedures are strictly bound to specific equipment parameters or alarm codes, we construct the \textit{Entity Graph} ($\mathcal{G}_{\text{E}}$). This subgraph serves as an entity-centric map, explicitly linking discrete domain entities to their governing Procedure Cards:
\begin{equation}
    \mathcal{G}_{\text{E}} = \{ (e, r, PC_i) \mid e \in \mathcal{V}_{Entity}\},
\end{equation}
where $\mathcal{V}_{Entity}$ represents the set of extracted entities (e.g., assets, parameters) and $r$ denotes the association relationship. This design prevents the semantic decoupling common in traditional chunking, ensuring that queries specifying unique proprietary constraints are routed directly to the correct equipment context.

\paragraph{Causal Graph.}
To address Condition-Dependent Relevance (C2), where the correct procedure depends on the underlying root cause rather than just surface symptoms, we build the \textit{Causal Graph} ($\mathcal{G}_{\text{C}}$). This graph models the non-linear condition-action logic, tracing the path from observable states to the resolving Procedure Card:
\begin{equation}
    \mathcal{G}_{\text{C}} = \{ (s_u, s_v) \mid s_u \xrightarrow{c} s_v \} \cup \{ (s, PC_i) \}
\end{equation}

Here, $\mathcal{V}_{State} = \{s_u, s_v, \dots\}$ represents system states (Symptoms, Causes), and the edges represent causal transitions. By traversing the path $(s_{start} \to \dots \to s_{end} \to PC_i)$, the system can simulate a diagnostic reasoning process, aligning retrieval with the operator's troubleshooting intent.

\paragraph{Flow Graph.}
To ensure Actionable Response Generation and preserve the procedural integrity of the response (C3), we generate a local \textit{Flow Graph} ($\mathcal{G}_{\text{F}}$) for each Procedure Card. Unlike flat text, this graph captures the precise topological structure required for safe execution:
\begin{equation}
    \mathcal{G}_{\text{F}}(PC_i) = (\mathcal{V}_{Step}, \mathcal{E}_{Flow}), \mathcal{E}_{Flow} \subseteq \mathcal{V}_{Step} \times \mathcal{V}_{Step}
\end{equation}

This structure $\mathcal{G}_{\text{F}}$ enables procedure segmentation that respects the inherent stepwise organization of each SOP, thereby avoiding the semantic fragmentation introduced by generic chunking strategies (C1). Each resulting segment remains tightly aligned with its corresponding $PC_i$, facilitating precise retrieval and providing operators with concise, reliable, step-by-step operational guidance.

\subsection{SOP Retrieval}
\sysname formalizes retrieval as an MoE-style gated inference pipeline. Adopting a "Coarse-to-Fine" strategy, the system emulates sparse gating by first pruning the search space via sparse activation and then adaptively routing queries to specialized reasoning experts for precise execution. It performs the following three specific steps.

\subsubsection{Procedure Card Layer Anchoring}
Given a query $Q$, the system first activates the Top-$K$ Procedure Cards to prune the search space. We calculate a global alignment score based on the semantic similarity between $Q$ and Procedure Cards:
\begin{equation}
    \mathcal{A}_{TopK} = \text{TopK}_{PC_i \in \mathcal{V}_{Global}} (\text{Sim}(Q, PC_i))
\end{equation}

This design reflects how operators typically search for SOPs by first scanning document titles, as these titles generally indicate the primary function or purpose of each procedure.

\subsubsection{LLM-Guided Routing}
To navigate the experts effectively, an Intention-Aware \textit{LLM Router} serves as the gating network. It dynamically modulates the contribution of each expert by assigning attention weights $\mathbf{w} = [w_E, w_C, w_F] = \text{Softmax}(\text{LLM}_{\text{Router}}(Q))$~\footnote{Details in Appendix~\ref{app:llm_router}, Appendix~\ref{app:case_llm_router}} based on the focus of the query:
\begin{itemize}[topsep=0pt, leftmargin=*, nosep]
    \item \textbf{Entity Focus ($w_E$):} For queries specifying equipment IDs or parameters.
    \item \textbf{Causal Focus ($w_C$):} For "Why" or diagnosis-oriented queries.
    \item \textbf{Flow Focus ($w_F$):} For "How" or execution-oriented queries.
\end{itemize}

\subsubsection{Subgraphs Reasoning}
The final relevance score $R(Q, S_i)$ is computed by aggregating the anchor score and the gated reasoning from the specialized experts:
\begin{align}
    R(Q, S_i) &= \lambda \cdot \text{Sim}(Q, PC_i) + (1-\lambda) \notag\\&\sum_{m \in \{E, C, F\}} w_m \cdot \mathcal{R}_m(Q, PC_i)
\end{align}
where $\mathcal{R}_m$ represents the specific scoring function for each subgraph:

\begin{enumerate}[topsep=0pt, leftmargin=*, nosep]
    \item \textbf{Entity Score ($\mathcal{R}_{\text{E}}$):} Measures the overlap of proprietary terms:
    \begin{align}
         \mathcal{R}_{\text{E}} &= \frac{1}{|\mathcal{V}_Q|} \sum_{e \in \mathcal{V}_Q}\Big( \alpha \cdot \mathbb{I}(e \in \mathcal{V}_{Entity}) \notag \\ 
        +& (1-\alpha) \cdot \max_{e' \in \mathcal{V}_{Entity}} \text{Sim}(e, e')\Big)
    \end{align}
    where $\mathcal{V}_{Q}$ denotes the set of entities extracted from $Q$, $\mathbb{I}(\cdot)$ is the indicator function for Exact Match, and $\alpha \in [0, 1]$ is a balancing weight.
    
    \item \textbf{Causal Score ($\mathcal{R}_{\text{C}}$):} Evaluates the causal validity. It identifies a start state $s$ matching $Q$ and measures the reachability to $PC_i$ by performing a $k$-hop neighborhood search on the causal graph $\mathcal{G}_{\text{C}}$:
    \begin{align}
        \mathcal{R}_{\text{C}} = \max_{s \in \mathcal{V}_{State}} \left( \text{Sim}(Q, s) \cdot \text{Path}(s \to PC_i) \right)
    \end{align}
    
    \item \textbf{Flow Score ($\mathcal{R}_{\text{F}}$):} Measures the semantic alignment with the executable steps within the local flow graph:
    \begin{equation}
        \mathcal{R}_{\text{F}} = \max_{p \in \mathcal{V}_{Step}(PC_i)} \text{Sim}(Q, p)
    \end{equation}
\end{enumerate}

\subsection{Structure-Aware Generation}
Once the optimal SOP is identified, the system retrieves its Flow Graph $\mathcal{G}_{\text{F}}(PC_i)$. This structured representation is linearized into a step-by-step prompt context (Figure~\ref{fig:soprag}). By constraining the LLM to this verified flow, we ensure the generation is precise and actionable, directly addressing the hallucination risks in industrial settings.

\section{SOP Benchmark Construction}
\label{sec:benchmark}

\begin{figure*}[t]
    \centering
    \includegraphics[width=0.9\linewidth]{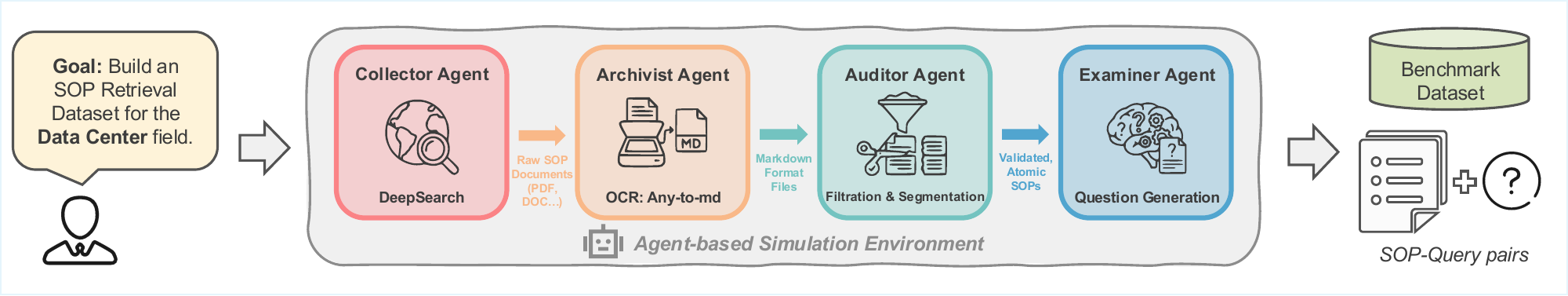}
    \caption{\textbf{The overview of the Agent-based Simulation Environment for automated dataset construction.} Given a specific domain, four specialized agents (\textit{Collector}, \textit{Archivist}, \textit{Auditor}, and \textit{Examiner}) mimic human expert workflows to automatically generate a high-quality benchmark dataset.}
    \label{fig:pipeline}
\end{figure*}

To rigorously evaluate retrieval algorithms in the domain of SOPs, we necessitate a high-quality, domain-specific benchmark, especially geared towards real-world industrial scenarios. Unlike general open-domain QA, SOP retrieval requires precise alignment between operational queries and procedural segments. 

To this end, we propose an Agent-based Simulation Environment for automated dataset construction. This pipeline simulates a real-world document processing lifecycle (data collection $\rightarrow$ processing $\rightarrow$ generation), ensuring both the diversity of the source data and logical validity of the queries.

\subsection{Agent Architecture}

As illustrated in Figure~\ref{fig:pipeline}, we conceptualize the data generation pipeline as a collaborative workflow involving four specialized agents: \textit{Collector}, \textit{Archivist}, \textit{Auditor}, and \textit{Examiner}. This multi-agent framework is designed to minimize human intervention while maximizing data quality.


\paragraph{Collector.} The foundation of the benchmark lies in the acquisition of raw, domain-specific documents.
The \textit{Collector} is instantiated as an LLM-powered agent equipped with DeepSearch capabilities. Unlike general web scrapers, it operates in a topic-driven manner. Given a specific industrial theme (e.g., "Data Center Maintenance"), the agent autonomously generates related search keywords and navigates the web to locate relevant resources. Its primary objective is to retrieve raw SOP documents that precisely align with the input topic, ensuring the benchmark covers targeted domains with high-quality source material.


\paragraph{Archivist.}
Raw data must be standardized and verified to serve as a valid retrieval corpus. To address the heterogeneity of file formats (e.g., PDFs, spreadsheets, and presentation slides), the \textit{Archivist} employs a layout-aware parsing agent powered by MinerU-2.5~\cite{niu2025mineru25decoupledvisionlanguagemodel}. Its primary role is document normalization: detecting document layouts and converting diverse raw files into a unified, clean Markdown format. This ensures subsequent agents can process the content as pure text, while simultaneously preserving the semantic integrity of the original structure (e.g., headers, tables).

\paragraph{Auditor.}
The \textit{Auditor} serves as the quality assurance layer, which performs two critical functions:

\begin{itemize}[topsep=0pt, leftmargin=*, nosep]
    \item \textbf{Filtration:} It analyzes the semantic content of the Markdown files to filter out non-SOP documents (e.g., general news, pure data tables) that may have bypassed the \textit{Collector}.
    
    \item \textbf{Segmentation:} Since a single document often contains multiple distinct procedures, the \textit{Auditor} splits multi-SOP files into atomic units. This ensures that the retrieval granularity is at the level of a single, executable procedure rather than a coarse-grained file.
\end{itemize}


\paragraph{Examiner.} Once the SOPs are validated and segmented into atomic units, we generate high-quality evaluation pairs to establish the benchmark ground truth. To construct realistic query-document pairs, we employ the \textit{Examiner}, a Question Generation Agent. For each validated SOP, the \textit{Examiner} adopts the persona of a frontline operator to simulate diverse real-world information needs. It is tasked with generating $N$ distinct questions per procedure, spanning a spectrum from simple keyword-based lookups to sophisticated, scenario-driven queries. This multi-level generation strategy ensures the benchmark accurately reflects the complex problem-solving requirements of industrial personnel in operational environments.

\subsection{Benchmark Statistics}
Leveraging this automated pipeline, we construct a comprehensive SOP benchmark comprising four domain-specific datasets to evaluate the performance of the proposed approach. Source documents for the Building Management and Airline Services domains are collected via the DeepSearch agent, while the \textit{Collector} in the Data Center and Liquid Cooling domains is substituted by manual curation to better reflect real-world industrial requirements. Table~\ref{tab:stats} summarizes the key statistics of the resulting dataset.

\begin{table}[ht]
\centering
\small
\setlength{\tabcolsep}{4.5pt}
\begin{tabular}{l l r r}
\toprule
\textbf{Domain} & \textbf{Source} & \textbf{\#Docs} & \textbf{\#Queries} \\
\midrule
Data Center         & Manual      & 414 & 250 \\
Liquid Cooling      & Manual      & 53  & 250 \\
Building Management & DeepSearch & 123 & 250 \\
Airline Services        & DeepSearch & 210 & 250 \\
\midrule
\textbf{Total}      & --          & \textbf{800} & \textbf{1000} \\
\bottomrule
\end{tabular}
\caption{\textbf{Detailed statistics of the SOP benchmark across four industrial domains.} For each domain, 50 atomic SOPs were randomly selected to generate 250 evaluation queries ($N=5$ queries per SOP).}
\label{tab:stats}
\end{table}

\begin{table*}[!t]
\centering
\resizebox{\textwidth}{!}{
\begin{tabular}{l
  |cccc
  |cccc
  |cccc
  |cccc}
\toprule
\multirow{2}{*}{\textbf{Method}} 
& \multicolumn{4}{c|}{\textbf{Data Center}}
& \multicolumn{4}{c|}{\textbf{Liquid Cooling}}
& \multicolumn{4}{c|}{\textbf{Building Management}}
& \multicolumn{4}{c}{\textbf{Airline Services}} \\
\cmidrule(lr){2-5} 
\cmidrule(lr){6-9} 
\cmidrule(lr){10-13} 
\cmidrule(lr){14-17}
    & \textbf{MRR} & \textbf{Acc@1} & \textbf{Acc@3} & \textbf{Acc@5}
    & \textbf{MRR} & \textbf{Acc@1} & \textbf{Acc@3} & \textbf{Acc@5}
    & \textbf{MRR} & \textbf{Acc@1} & \textbf{Acc@3} & \textbf{Acc@5}
    & \textbf{MRR} & \textbf{Acc@1} & \textbf{Acc@3} & \textbf{Acc@5} \\
\midrule
BM25        & 0.39 & 0.28 & 0.48 & 0.55  
            & 0.58 & 0.45 & 0.60 & 0.70  
            & 0.67 & 0.58 & 0.73 & 0.82  
            & 0.69 & 0.58 & 0.79 & 0.83 \\
OpenAI Embed  & 0.39 & 0.30 & 0.47 & 0.53  
            & 0.54 & 0.38 & 0.69 & 0.75  
            & 0.68 & 0.63 & 0.82 & 0.86  
            & 0.75 & 0.67 & 0.83 & 0.86 \\
\midrule
GraphRAG   & 0.27 & 0.16 & 0.35 & 0.46  
            & 0.28  & 0.14 & 0.36  & 0.53  
            & 0.24  & 0.10 & 0.28  & 0.57  
            & 0.40  & 0.26 & 0.49  & 0.67 \\
LightRAG   & 0.29 & 0.16 & 0.40 & 0.55  
            & 0.32 & 0.14 & 0.42 & 0.70  
            & 0.31 & 0.15 & 0.41 & 0.63  
            & 0.43 & 0.29 & 0.54 & 0.71 \\
HippoRAG 2  & 0.31 & 0.20 & 0.42 & 0.54  
            & 0.55 & 0.42 & 0.67 & 0.76  
            & 0.64 & 0.56 & 0.70 & 0.79  
            & 0.63 & 0.52 & 0.70 & 0.86 \\
\textbf{\sysname}
            & \textbf{0.49} & \textbf{0.40} & \textbf{0.57} & \textbf{0.64} 
            & \textbf{0.59} & \textbf{0.46} & \textbf{0.71} & \textbf{0.78} 
            & \textbf{0.73} & \textbf{0.63} & \textbf{0.85} & \textbf{0.91} 
            & \textbf{0.76} & \textbf{0.64} & \textbf{0.89} & \textbf{0.93} \\
\bottomrule
\end{tabular}
}
\caption{\textbf{Retrieval performance across four domain-specific datasets, measured by MRR and Acc@K.}}
\label{tab:retrieval_main}
\end{table*}



\section{Experiments and Results}
\label{sec:experiments}

To evaluate the effectiveness of \sysname, we conducted extensive experiments on the domain-specific SOP benchmark constructed in Section \ref{sec:benchmark}. The benchmark covers four diverse industrial fields: Data Center, Liquid Cooling, Building Management, and Airline Services, ensuring a comprehensive assessment of \sysname's adaptability to different operational contexts.

\begin{figure}[h]
    \centering
    \includegraphics[width=0.9\linewidth]{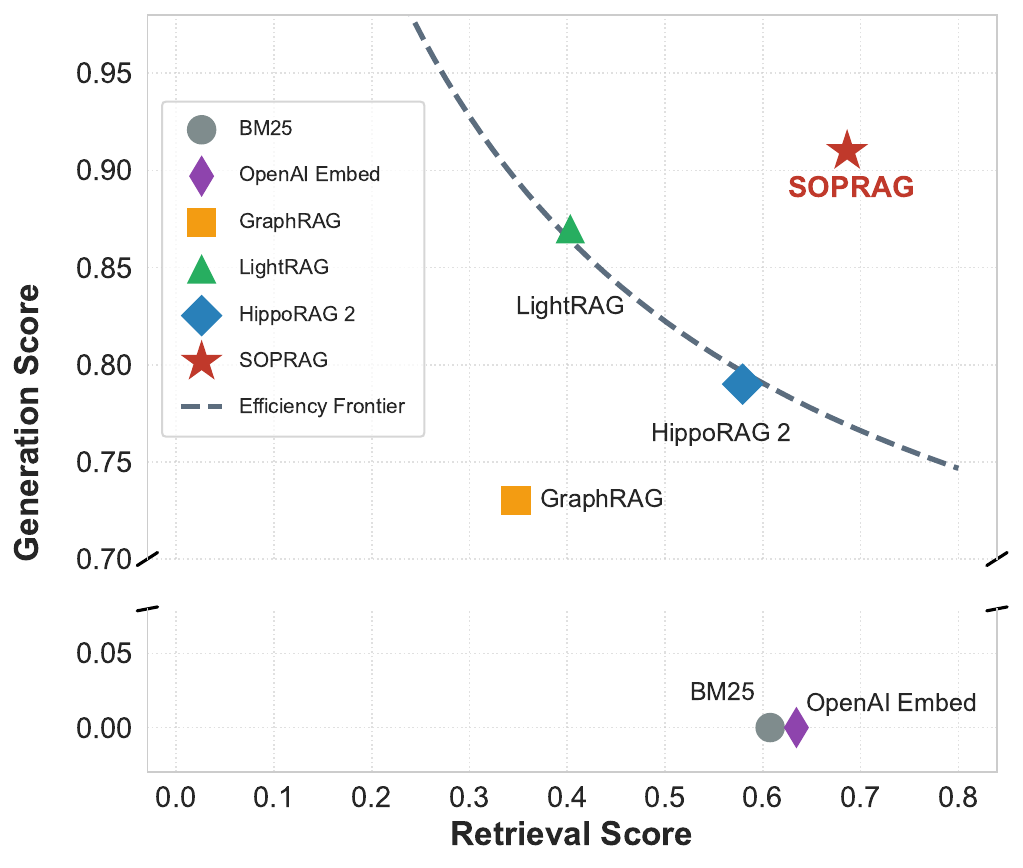}
    \caption{\textbf{Retrieval vs. Generation Trade-off.} Points represent mean scores of all evaluation metrics across four domain-specific datasets. The dashed line indicates the Efficiency Frontier of existing methods. \sysname noticeably transcends this boundary, demonstrating an optimal balance between retrieval and generation.}
    \label{fig:tradeoff}
\end{figure}

\subsection{Experimental Settings}

\paragraph{Baselines.}
We compare our method against three categories of strong baselines:
\begin{itemize}[topsep=0pt, leftmargin=*, nosep]
    \item \textbf{Lexical Retrieval (BM25):} A standard probabilistic retrieval method based on exact term matching~\citep{Robertson2009bm25}.
    \item \textbf{Dense Retrieval (OpenAI Embed):} Utilizing the \texttt{text-embedding-3-small} model to perform semantic search in a dense vector space.
    \item \textbf{Graph-based RAG:} We primarily compare with the current state-of-the-art graph-based RAG systems, including \textbf{GraphRAG}~\citep{edge2024local}, \textbf{LightRAG}~\citep{guo-etal-2025-lightrag}, and \textbf{HippoRAG2}~\citep{gutiérrez2025ragmemorynonparametriccontinual}.
\end{itemize}
For all graph-based methods (including ours) and generation tasks, we utilize GPT-4o as the backbone LLM to ensure fair comparison under consistent model capabilities.

\subsubsection{Evaluation Metrics}
We assess performance across two dimensions:
\begin{itemize}[topsep=0pt, leftmargin=*, nosep]
    \item \textbf{Retrieval Quality:} We report MRR (Mean Reciprocal Rank) and Accuracy@K (K=1, 3, 5) to measure the system's ability to locate the correct SOP.
    \item \textbf{Generation Quality:} We employ the RAGAS~\citep{ragas2024} framework to evaluate \textit{Faithfulness}, \textit{Answer Relevancy}, and \textit{Context Precision}. Additionally, we introduce a domain-specific metric, \textit{SOP Quality Score}~\footnote{Details in Appendix~\ref{app:SOP_Quality_Score}}, which uses an LLM judge to evaluate the structural completeness and executability of the generated steps.
\end{itemize}

\begin{table*}[!t]
\centering
\resizebox{\textwidth}{!}{
\begin{tabular}{l
cccc
cccc
cccc
cccc}
\toprule
\multirow{2}{*}{\textbf{Method}} 
& \multicolumn{4}{c}{\textbf{Data Center}}
& \multicolumn{4}{c}{\textbf{Liquid Cooling}}
& \multicolumn{4}{c}{\textbf{Building Management}}
& \multicolumn{4}{c}{\textbf{Airline Services}} \\
\cmidrule(lr){2-5} \cmidrule(lr){6-9} \cmidrule(lr){10-13} \cmidrule(lr){14-17}
    & faith. & ans. & ctx. & sop. 
    & faith. & ans. & ctx. & sop.
    & faith. & ans. & ctx. & sop.
    & faith. & ans. & ctx. & sop. \\
\midrule
GraphRAG    & 0.47 & \underline{0.97} & \underline{0.62} & 0.79  & 0.44 & 0.92 & 0.66 & 0.64  & 0.49 & 0.91 & 0.77 & 0.79  & 0.65 & \underline{0.96} & \underline{0.90} & 0.71 \\
LightRAG    & 0.70 & \textbf{0.98} & 0.54 & \underline{0.93}  & 0.79 & \textbf{0.97} & \textbf{0.93} & \underline{0.82}  & 0.83 & \textbf{0.97} & \underline{0.82} & \underline{0.93}  & 0.86 & \underline{0.96} & \textbf{0.93} & \underline{0.92} \\
HippoRAG 2   & \underline{0.72} & 0.83 & 0.52 & 0.66  & \textbf{0.90} & 0.86 & 0.85 & 0.62  & \textbf{0.97} & 0.93 & 0.78 & \underline{0.93}  & \textbf{0.89} & 0.77 & 0.88 & 0.51 \\
\textbf{\sysname}
            & \textbf{0.77} & 0.96 & \textbf{0.66} & \textbf{1.00}  
            & \underline{0.82} & \textbf{0.97} & \underline{0.92} & \textbf{0.99} 
            & \underline{0.88} & \textbf{0.97} & \textbf{0.91} & \textbf{0.99}
            & \underline{0.88} & \textbf{0.97} & 0.84 & \textbf{0.98} \\
\bottomrule
\end{tabular}
}
\caption{\textbf{Generation quality results across four domain-specific datasets, evaluated in terms of Faithfulness (faith.), Answer Relevancy (ans.), Context Precision (ctx.), and SOP Quality Score (sop.).}}
\label{tab:generation_main}
\end{table*}

\subsection{Main Results}

\subsubsection{Retrieval Performance}
Table \ref{tab:retrieval_main} presents the retrieval performance across four domains. 
\sysname consistently outperforms all baselines in terms of MRR and Accuracy.

First, \sysname demonstrates distinct retrieval advantages over traditional lexical, dense, and general graph-based baselines. It consistently achieves the highest MRR and Accuracy, reaching a peak MRR of 0.76 and Acc@5 of 0.93 in the Airline Services domain. This superior performance stems from the hierarchical integration of Procedure Cards and multi-view subgraphs, which effectively resolves the "Proprietary Structure" challenge that often causes standard semantic models to conflate similar but contextually distinct procedures.

Second, the stark performance gap between \sysname and general graph-based methods highlights the necessity of domain-specific structural modeling. While general frameworks prioritize global community summarization, \sysname utilizes a Causal Graph to navigate scenario-dependent relevance and a Flow Graph to align sequential execution steps. By leveraging an LLM-guided routing mechanism to dynamically weight these experts, the system moves beyond simple semantic matching to perform intent-aware topological reasoning, ensuring high precision even in complex, low-fault-tolerance industrial environments.

\subsubsection{Generation Performance}
We further evaluate the generation quality, comparing our method against other graph-based approaches. As shown in Table \ref{tab:generation_main}, \sysname consistently achieves the highest SOP Quality Score.

By explicitly retrieving the \textit{Flow Graph} and linearizing it into the prompt, our method ensures the generated response follows a verified sequence. This results in a perfect SOP Quality Score of 1.0 in the Data Center task, indicating that the outputs are directly executable and structurally sound compared to the more hallucination-prone baselines.

\subsection{Ablation Study}

To investigate the contribution of each component in our framework, we conducted an ablation study. We analyzed the impact of removing the Procedure Card (PC), individual subgraph experts (Entity, Causal, Flow), and the LLM Routing mechanism. Based on the results in Table \ref{tab:ablation}, we highlight the following key findings:

\begin{table}[!h]
\centering
\resizebox{\linewidth}{!}{
\begin{tabular}{l ccc ccc}
\toprule
\multirow{2}{*}{\textbf{Variant}} 
& \multicolumn{3}{c}{\textbf{Data Center}} 
& \multicolumn{3}{c}{\textbf{Airline Services}} \\
\cmidrule(lr){2-4}
\cmidrule(lr){5-7}
& \textbf{MRR} & \textbf{Acc@1} & \textbf{Acc@5}
& \textbf{MRR} & \textbf{Acc@1} & \textbf{Acc@5} \\
\midrule
w/o PC       & 0.47 & 0.36 & 0.64 & 0.73 & 0.61 & 0.90 \\
w/o Entity   & 0.41 & 0.31 & 0.56 & \textbf{0.76} & \textbf{0.66} & 0.90 \\
w/o Causal   & 0.44 & 0.36 & 0.60 & 0.68 & 0.55 & 0.89 \\
w/o Flow     & 0.39 & 0.28 & 0.58 & 0.68 & 0.56 & 0.88 \\
\midrule
Full (Avg)   & 0.45 & 0.34 & 0.63 & 0.75 & 0.65 & 0.89 \\
\textbf{Routing} & \textbf{0.49} & \textbf{0.40} & \textbf{0.64} & \textbf{0.76} & \textbf{0.64} & \textbf{0.93} \\
\bottomrule
\end{tabular}
}
\caption{Ablation study of different components in \sysname.}
\label{tab:ablation}
\end{table}

\noindent\textbf{PC Layer Efficiency:} The Procedure Card acts as a vital hierarchical filter. Its removal (w/o PC) leads to a consistent drop in both MRR and Accuracy across both the Data Center and Airline Services domains, confirming its role in pruning the search space before detailed reasoning.

\noindent\textbf{Subgraph Synergy:} Each subgraph expert addresses specific SOP complexities. The Flow Graph is most critical in the Data Center dataset, where its removal (w/o Flow) results in the largest performance drop (MRR to 0.39). The Causal Graph is similarly essential in the Airline Services dataset, where its absence (w/o Causal) reduces MRR from 0.76 to 0.68.

\noindent\textbf{Routing Optimization:} The dynamic Routing mechanism outperforms static averaging (Full Avg) by adaptively weighting subgraphs based on intent. It boosts Data Center Acc@1 from 0.34 to 0.40 and achieves a peak Acc@5 of 0.93 in Airline Services.

\section{Conclusion}
In this work, we addressed the critical yet underserved problem of SOP retrieval in industrial settings. We identified three primary hurdles, proprietary structure, condition-dependency, and the requirement for actionable responses, which render traditional RAG methods insufficient. To overcome these, we introduced \sysname, a structure-aware framework that emulates human cognitive patterns through a "Retrieval-based Mixture-of-Experts" architecture. By integrating hierarchical Procedure Cards with specialized subgraph experts for entities, causal logic, and sequential flows, our system provides precise, streamlined support to operators while mitigating the risks of hallucination. Furthermore, our automated agentic pipeline provides a scalable solution for constructing high-quality industrial benchmarks. Experimental results confirm that explicitly modeling the procedural "shape" of knowledge is essential for industrial troubleshooting, with our approach consistently achieving superior retrieval precision and generation quality scores across diverse operational contexts.

\section*{Limitations}
Despite the superior performance of \sysname, several limitations remain to be addressed in future research. 
First, the framework primarily processes layout-aware Markdown text and table but lacks the integration of multimodal information, such as industrial diagrams or safety icons, which are common in SOPs. 
Second, the fidelity of the specialized Entity, Causal, and Flow graphs is inherently dependent on the reasoning capabilities of the LLM used during the construction phase; although the system supports a "construct-offline, infer-online" paradigm to enable the use of smaller models for real-time inference, the specific efficacy and trade-offs of such small-model reasoning require further empirical validation. 
Finally, while our automated agentic pipeline efficiently generates diverse evaluation queries , these queries are currently synthetic and LLM-generated. Future work could incorporate expert-crafted questions and human-in-the-loop validation to construct even higher-quality benchmarks that more accurately reflect the nuanced information needs of frontline operators.

\bibliography{custom}

\newpage
\appendix


\section{System Prompts for \sysname Components}
\label{app:system_prompt}
This section details the system prompts used for multi-view subgraph construction and the LLM-Guided Routing mechanism.

\subsection{Multi-view Graph Construction Prompts}
To address the challenges of proprietary structures (C1), condition-dependency (C2), and actionable execution (C3), we use the following specialized prompts to extract structured knowledge from raw SOP documents:

\begin{promptbox}{Entity Graph Construction ($\mathcal{G}_E$)}
You are an expert at extracting key entities from SOP (Standard Operating Procedure) documents.

Focus on identifying:

- Alarms (with codes like A01, ALARM\-123, etc.)

- Parameters (temperature, pressure, flow rate, etc.)

- Assets/Equipment (pumps, valves, sensors, etc.)

- Roles (operator, supervisor, engineer, etc.)

- Other important named entities relevant to SOPs.

Noted: Determine its attributes based on the context in the SOP. Avoid including single numbers or letters.
\end{promptbox}

\begin{promptbox}{Causal Graph Construction ($\mathcal{G}_C$)}
You are an expert at identifying cause-effect relationships in SOP (Standard Operating Procedure) documents.

Extract causal relationships such as:

- What causes what (X causes Y)

- What prevents what (X prevents Y)

- What enables what (X enables Y)

- What mitigates what (X mitigates Y)
\end{promptbox}

\begin{promptbox}{Flow Graph Construction ($\mathcal{G}_F$)}
You are an expert at extracting procedural flows from SOP (Standard Operating Procedure) documents.

Extract the step-by-step procedure as a flowchart structure with:

- Action steps (things to do)

- Condition/decision steps (checks, branches)

- Connections between steps

Noted: Do NOT create entry/exit nodes - focus only on the actual procedural steps.
\end{promptbox}

\subsection{LLM-Guided Routing Mechanism Prompt}
\label{app:llm_router}
The Intention-Aware LLM Router employs the following prompt to characterize user intent and extract key entities, denoted as $\text{LLM}_{\text{Router}}(Q)$, which subsequently informs the dynamic modulation of the routing weights $w = [w_E, w_C, w_F]$.

\begin{promptbox}{LLM-Guided Routing Mechanism Prompt}

You are an expert at understanding user queries about Standard Operating Procedures.

Your task is to analyze the query and determine:

1. The user's intent (what is the operator trying to achieve?)

2. Key entities mentioned

3. Which subgraph experts should be prioritized:

   - Entity graph: Focus on specific alarms, parameters, equipment
   
   - Causal graph: Focus on cause-effect, why something happens
   
   - Flow graph: Focus on step-by-step procedures, how to do something

\end{promptbox}

\section{SOP Quality Score}
\label{app:SOP_Quality_Score}

To better evaluate the practical utility of generated instructions in industrial scenarios, we introduce the \textit{SOP Quality Score}, an LLM-as-Judge metric integrated into the RAGAS framework. Unlike general fluency or faithfulness scores, this metric specifically assesses whether the output satisfies the requirements of Actionable Response Generation (C3) in SOP Retrieval. We use GPT-4o as the evaluator to score the generated response on a scale of 0 to 1 based on its structural alignment with the source SOP and its technical executability.

The evaluation logic is encapsulated in the following system prompt:
\begin{promptbox}{System Prompt for SOP Quality Score}
You are evaluating the quality of an answer generated by a SOP (Standard Operating Procedure) RAG system.

The ideal answer for a SOP question should:

1. Provide clear step-by-step instructions on how to perform the task

2. Be precise and specific (avoid vague or ambiguous language, align with the source SOP)

3. Be concise (include only necessary information, no redundancy)

4. Be clear and easy to understand (well-structured, logical flow)

Evaluate the given answer based on these criteria and provide:

- has\_step\_by\_step: 1 if the answer provides step-by-step instructions, 0 otherwise

- is\_precise: 1 if the answer is precise and specific, 0 if it's vague

- is\_concise: 1 if the answer is concise, 0 if it's verbose or contains unnecessary information

- is\_clear: 1 if the answer is clear and easy to understand, 0 if it's confusing

- score: Overall quality score (0.0 to 1.0) calculated as the average of the four criteria

- reasoning: Brief explanation of your evaluation
\end{promptbox}

\section{Ethical Considerations}

The benchmark datasets introduced in this work are developed strictly for scientific research purposes to evaluate the performance of various RAG paradigms in industrial SOP retrieval tasks. 
The Data Center and Liquid Cooling datasets are derived from authentic industrial scenarios. Consequently, they remain confidential and will not be publicly disclosed to protect proprietary information. 
The Building Management and Airline Services datasets were constructed based on publicly available SOPs retrieved from the web using our DeepSearch agent. To support transparency and academic reproducibility, we can provide the source links and details of our agentic workflow for these two datasets. 
All data utilization is aimed at enhancing operational safety and efficiency in real-world industrial environments.

\section{Analysis of Gating Mechanism Behavior}
\label{app:router}

To further investigate how \sysname adaptively routes queries to specialized experts, we visualize the Kernel Density Estimation (KDE) of the attention weights $w = [w_E, w_C, w_F]$ across four industrial domains. As shown in Figure \ref{fig:multi_kde}, the routing patterns exhibit several key characteristics:

\begin{figure}[h]
    \centering
    \includegraphics[width=1.0\linewidth]{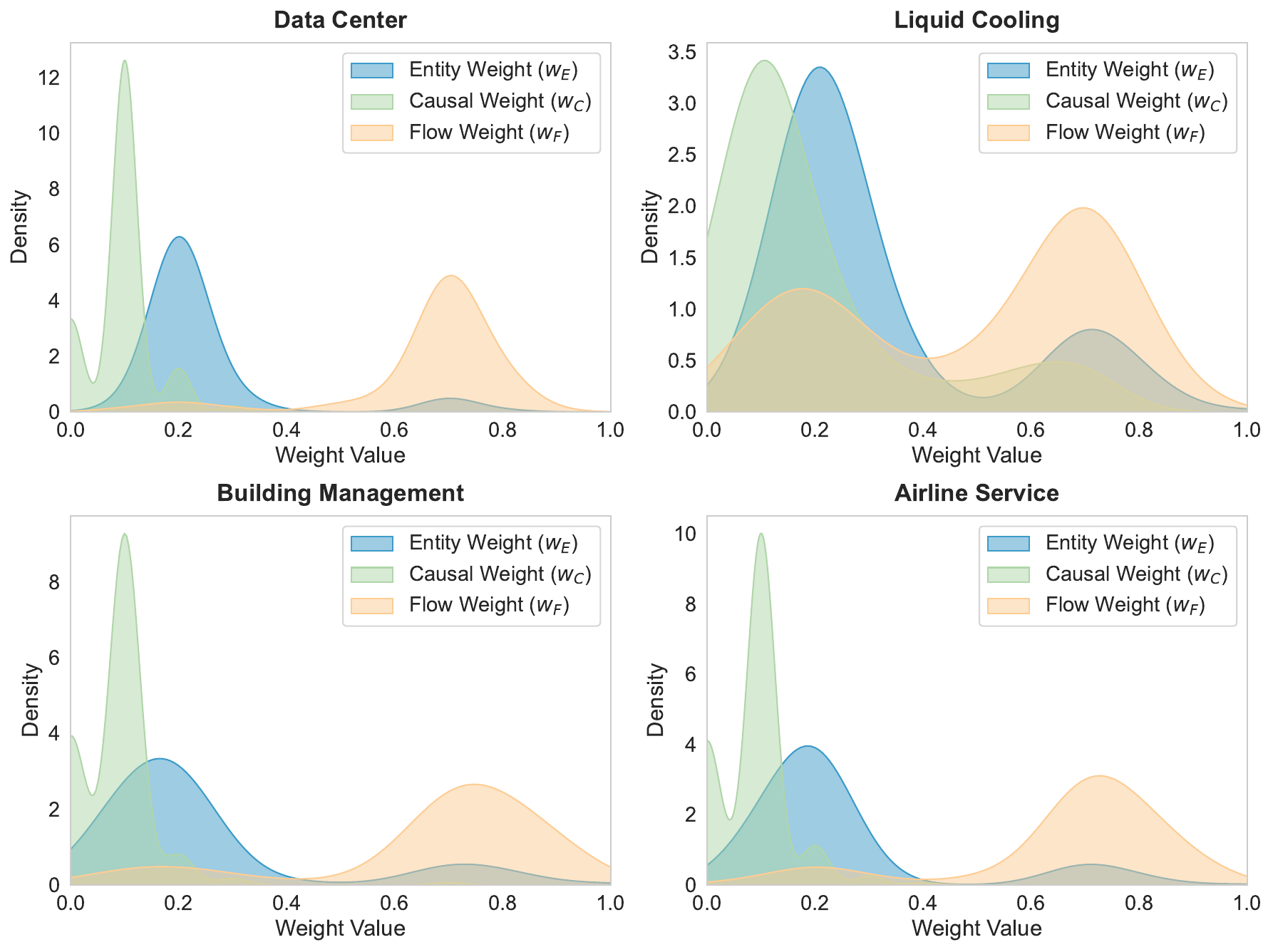}
    \caption{\textbf{Kernel Density Estimation (KDE) of gating weights ($w_E, w_C, w_F$) across four industrial domains, based on 250 evaluation queries per domain.}}
    \label{fig:multi_kde}
\end{figure}

\begin{itemize}
    \item \textbf{Dominance of Flow Logic ($w_F$)}: Across all domains, the density of $w_F$ (orange) consistently peaks in the $[0.6, 0.8]$ range. This empirical evidence confirms that the router prioritizes procedural execution steps, directly addressing the requirement for Actionable Response Generation (C3) in industrial settings.

    \item \textbf{Specialized Sparse Activation}: The $w_C$ (green) and $w_E$ (blue) weights exhibit sharp peaks at lower values ($\approx 0.1$ and $0.2$, respectively), indicating a sparse activation strategy. These experts are only "called upon" with high weights when the router identifies specific causal symptoms or proprietary entity constraints within the operator's query.

    \item \textbf{Cross-Domain Adaptation}: The activation profile adaptively shifts based on domain characteristics. For instance, the Data Center domain shows the most concentrated $w_F$ peak, reflecting its highly standardized procedural nature. In contrast, the Liquid Cooling domain demonstrates a unique profile where $w_C$ receives a significantly higher frequency of high-score assignments (peaking near 0.7) compared to other domains. This shift is consistent with the increased necessity for causal reasoning and diagnostic logic in complex thermal management scenarios, where identifying the root cause of a system state is often a prerequisite for safe procedural execution.
\end{itemize}

This distribution validates that the LLM-Guided Router performs intent-aware topological reasoning rather than simple semantic matching, ensuring high precision in low-fault-tolerance environments.

\section{Analysis of Computational Cost and Efficiency}

\begin{figure}[h]
    \centering
    \includegraphics[width=1.0\linewidth]{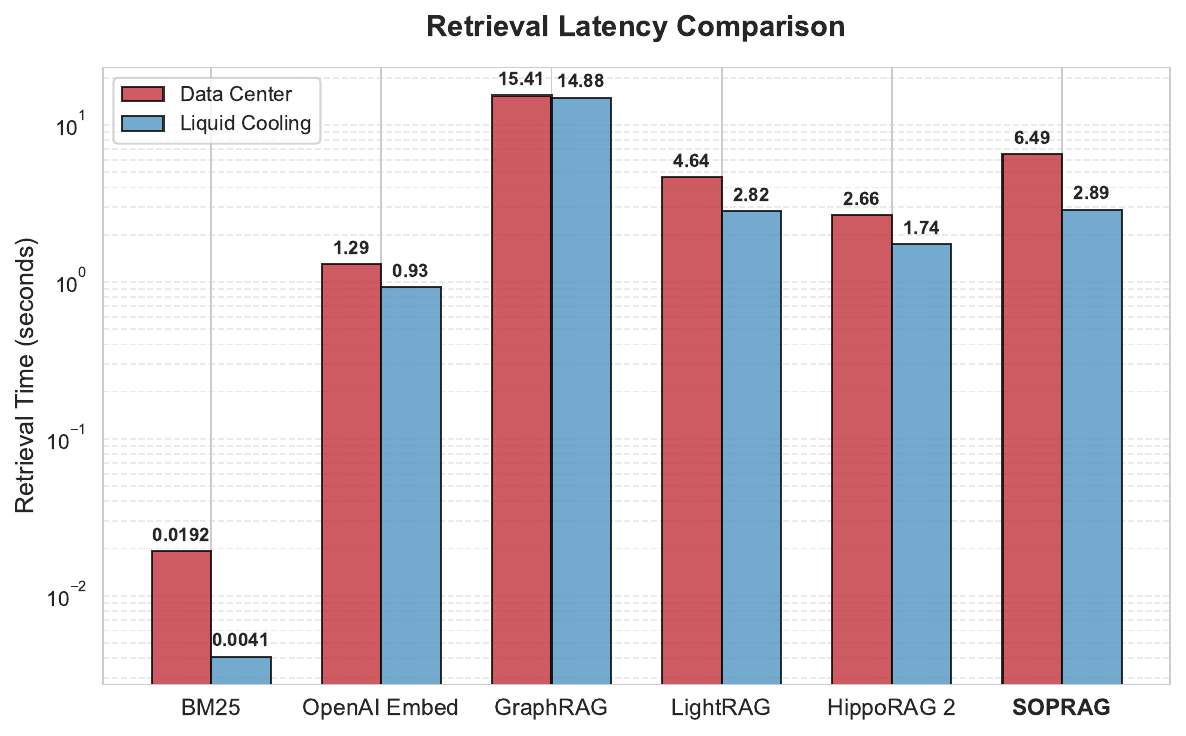}
    \caption{\textbf{Retrieval Latency.}}
    \label{fig:retrieval_time}
\end{figure}

To evaluate the practical feasibility of \sysname, we conduct a comparative analysis of retrieval latency against various baselines. As illustrated in Figure~\ref{fig:retrieval_time}, while \sysname exhibits a slightly higher retrieval time compared to lightweight lexical or dense retrieval methods (e.g., BM25 and OpenAI Embed), it remains highly competitive within the landscape of state-of-the-art graph-based RAG systems.

Specifically, in the Liquid Cooling domain, \sysname achieves a retrieval time of approximately 2.89 seconds, which is significantly faster than the standard GraphRAG (over 14 seconds) and comparable to LightRAG. Although the complexity of the multi-view gating mechanism leads to a slight overhead compared to HippoRAG 2, the trade-off is balanced by the significant gains in retrieval accuracy and structural consistency. In mission-critical industrial troubleshooting, where high-precision guidance is paramount, a response time of under 7 seconds effectively bridges the gap between manual document search and real-time operator assistance.

\section{Incremental Graph Construction and Scalability}
To ensure the scalability of \sysname in dynamic industrial environments, our framework supports efficient incremental updates as new procedures are introduced. Upon the acquisition of a new SOP, the system performs the standard multi-view extraction process to construct its local Entity, Causal, and Flow subgraphs.

The integration process follows a decoupled logic:
\begin{itemize}
    \item Atomic Flow Preservation: Since the Flow Graph ($\mathcal{G}_F$) is designed to preserve the specific sequential integrity of an atomic procedure, it is stored as a standalone structured unit linked to its corresponding Procedure Card.
    \item Global Knowledge Merging: The newly extracted Entity Graph ($\mathcal{G}_E$) and Causal Graph ($\mathcal{G}_C$) are integrated into the existing global knowledge base through a similarity-based merging mechanism. Specifically, nodes representing identical or highly similar equipment entities, parameters, or system states are consolidated to maintain cross-procedural connectivity
\end{itemize}

This incremental design allows \sysname to expand its procedural intelligence with minimal computational overhead, avoiding the need for global graph re-construction while ensuring the retriever remains adaptively updated with emerging operational contexts.

\section{Agent-Integrated Operational Execution}
\label{app:agent-exe}

Beyond static retrieval, the structured nature of \sysname's Flow Graph ($\mathcal{G}_F$) provides a robust foundation for integration with autonomous LLM Agents. By mapping individual nodes within the Flow Graph to specific agentic tools, the system transforms procedural instructions into executable workflows.

In this integrated framework, the Structure-Aware Generation does not merely output text but serves as a controller for specialized agents. For instance, a retrieved step requiring "notifying the team" can trigger an Email Agent to automatically draft the message content and populate recipient fields based on the SOP's context. Similarly, steps involving "calling a supervisor" can be facilitated through automated dialing and logging tools. 

This synergy between \sysname and agentic tools significantly enhances operational efficiency by reducing manual data entry and cognitive load for frontline personnel. By maintaining a human-in-the-loop approach, where the agent prepares the execution environment and the operator provides the final authorization, the system ensures high-precision execution while mitigating the risks of manual oversight in mission-critical industrial environments.

\begin{figure}[h]
    \centering
    \includegraphics[width=1.0\linewidth]{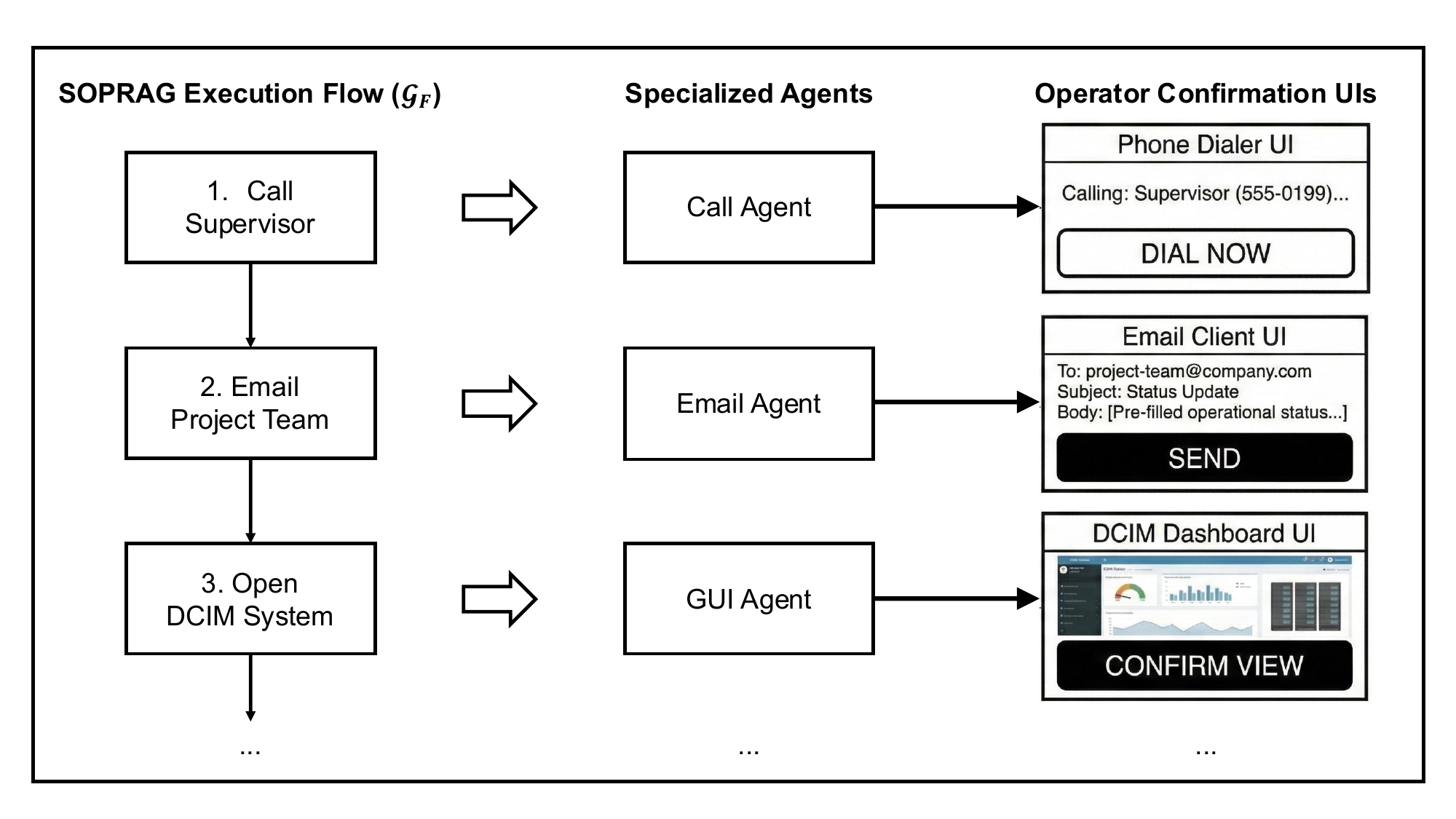}
    \caption{\textbf{\sysname Agent-Integrated Operational Execution.} The structured flow triggers specialized agents, which prepare actions for operator confirmation, streamlining complex multi-step procedures.}
    \label{fig:agentic_exe}
\end{figure}

\section{Case Studies of LLM Router Output}
\label{app:case_llm_router}
This section provides concrete examples of the Intention-Aware LLM Router's weight assignments across different query types. These cases illustrate how the gating mechanism dynamically modulates attention weights $w = [w_E, w_C, w_F]$ to align retrieval with the operator's specific information needs.

\begin{promptbox}{Case 1: Entity-Oriented Query (Asset Monitoring)}
\textbf{Query:} "How can I view and monitor the real-time operating status of HVAC equipment in the chilled water and cooling water systems?" \\
\textbf{Router Weights:} $w_E: 0.7, w_C: 0.0, w_F: 0.3$
\end{promptbox}

\textbf{Analysis:} This query focuses on specific proprietary assets ("HVAC equipment", "chilled water systems"). The Router assigns a high Entity Weight ($w_E$) to ensure precise anchoring to the relevant equipment nodes in the knowledge graph, minimizing semantic interference from unrelated systems.

\begin{promptbox}{Case 2: Causal-Oriented Query (Diagnostic Reasoning)}
\textbf{Query:} "How can increasing air and water temperatures in a chilled water system improve cooling efficiency and sustainability?" \\
\textbf{Router Weights:} $w_E: 0.2, w_C: 0.7, w_F: 0.1$
\end{promptbox}

\textbf{Analysis:} This query seeks an understanding of the underlying relationship between system states and performance outcomes. The high Causal Weight ($w_C$) directs the system to traverse the Causal Graph ($\mathcal{G}_C$), facilitating diagnostic reasoning rather than simple step retrieval.

\begin{promptbox}{Case 3: Flow-Oriented Query (Actionable Execution)}
\textbf{Query:} "What is the standard procedure for recovering from an aircraft upset or unusual attitude in flight?" \\
\textbf{Router Weights:} $w_E: 0.1, w_C: 0.1, w_F: 0.8$
\end{promptbox}

\textbf{Analysis:} The operator explicitly requests a "standard procedure," which is inherently sequential and actionable. The Router correctly allocates the majority of the weight to the Flow Graph expert ($w_F$), ensuring the response is structured as a precise, step-by-step workflow for safe execution.

\end{document}